# ClassWise-CRF: Category-Specific Fusion for Enhanced Semantic Segmentation of Remote Sensing Imagery

Qinfeng Zhu[a,b], Yunxi Jiang[c], Lei Fan[a,d,1]

[a] Department of Civil Engineering, Xi'an Jiaotong-Liverpool University, Suzhou, 215123, China

[b] Department of Computer Science, University of Liverpool, Liverpool, L69 3BX, UK

[c] Department of Physics, University of Liverpool, Liverpool, L69 7ZE, UK

[d] Design School Intelligent Built Environment Research Centre, Xi'an Jiaotong-Liverpool University, Suzhou, 215123, China

**Abstract**: With the continuous development of visual models such as Convolutional Neural Networks, Vision Transformers, and Vision Mamba, the capabilities of neural networks in semantic segmentation of remote sensing images have seen significant progress. However, these networks exhibit varying performance across different semantic categories, making it challenging to find a single network architecture that excels in all categories. To address this, we propose a result-level category-specific fusion architecture called ClassWise-CRF. This architecture employs a two-stage process: first, it selects expert networks that perform well in specific categories from a pool of candidate networks using a greedy algorithm; second, it integrates the segmentation predictions of these selected networks by adaptively weighting their contributions based on their segmentation performance in each category. Inspired by Conditional Random Field (CRF), the ClassWise-CRF architecture treats the segmentation predictions from multiple networks as confidence vector fields. It leverages segmentation metrics (such as Intersection over Union) from the validation set as priors and employs an exponential weighting strategy to fuse the category-specific confidence scores

[1] Corresponding author.

Email addresses: Lei.Fan@xjtlu.edu.cn (L. Fan)

predicted by each network. This fusion method dynamically adjusts the weights of each network for different categories, achieving category-specific optimization. Building on this, the architecture further optimizes the fused results using unary and pairwise potentials in CRF to ensure spatial consistency and boundary accuracy. To validate the effectiveness of ClassWise-CRF, we conducted experiments on two remote sensing datasets, LoveDA and Vaihingen, using eight classic and advanced semantic segmentation networks. The results show that the ClassWise-CRF architecture significantly improves segmentation performance: on the LoveDA dataset, the mean Intersection over Union (mIoU) metric increased by 1.00% on the validation set and by 0.68% on the test set; on the Vaihingen dataset, the mIoU improved by 0.87% on the validation set and by 0.91% on the test set. These results fully demonstrate the effectiveness and generality of the ClassWise-CRF architecture in semantic segmentation of remote sensing images. The full code is available at https://github.com/zhuqinfeng1999/ClassWise-CRF.

## 1 Introduction

Semantic segmentation is a pivotal research domain within computer vision, with the objective of precisely assigning each pixel in an image to a predefined semantic category [1]. A key application of this technique is the semantic segmentation of remote sensing images [2, 3], which aims to automate the identification and analysis of land cover types and land use patterns [4]. With the rapid advancement of remote sensing technology, the spatial resolution and coverage of satellite and aerial imaging systems have markedly improved, delivering vast and detailed datasets that support applications including urban planning, agricultural monitoring, and disaster assessment [5]. Nevertheless, the intricate characteristics of remote sensing images [6], such as the presence of multi-scale objects, diverse background environments, and dynamic lighting conditions, pose challenges to the accuracy and robustness of semantic segmentation models [7].

In recent years, groundbreaking progress in deep learning techniques have significantly boosted the development of various semantic segmentation methods for remote sensing images, making deep learning the predominant methodology for the segmentation task[2]. As a cornerstone of deep learning in image processing, Convolutional Neural Networks (CNNs) leverage their localized receptive fields and hierarchical feature extraction capabilities to excel at capturing local textures and structural details in remote sensing imagery [8]. For instance, Fully Convolutional Networks (FCNs) [9] and their variants such as U-Net [10] and the DeepLab series [11] have achieved notable success in remote sensing image segmentation by accurately modeling fine details like building edges and road textures [12]. However, CNNs exhibit inherent limitations in processing global contextual information [13]. Given the complex spatial distribution of land cover categories in remote sensing images, which frequently involve cross-scale dependencies, an exclusive reliance on local features may result in inadequate segmentation accuracy for large-scale objects, such as expansive water bodies or forests [14].

To overcome the deficiencies of CNNs in global modeling, models based on self-attention mechanisms, such as Vision Transformers (ViTs) [15, 16], have emerged as a promising alternative. By segmenting images into patches and performing global modeling, ViTs effectively capture long-range dependencies, thereby enhancing the comprehension of overall semantic information in remote sensing images [8]. For example, in identifying large-scale water bodies or vegetation coverage, ViTs often outperform CNNs [17]. Nevertheless, ViTs may fall short of CNNs in handling fine-grained local details, particularly in high-resolution remote sensing images, where the precise segmentation of building boundaries or road features imposes greater demands on models [18].

Subsequently, more novel architectures seek to integrate the strengths of CNNs and Transformers, striving for a balanced extraction of local and global features. For instance, the Swin Transformer [19] employs a hierarchical window-based self-attention mechanism to bolster local feature perception while preserving global modeling capabilities. Similarly, ConvNeXt [20] modernizes the CNN framework, achieving computational efficiency and performance levels comparable to those of Transformers. More recently, novel Vision Mamba [21] architectures realize global attention with linear computational complexity, and have demonstrated superior performance over ViT-based approaches in high-resolution semantic segmentation tasks [18, 22]. However, despite their distinct advantages, achieving consistently optimal segmentation performance across all semantic categories remains a persistent challenge.

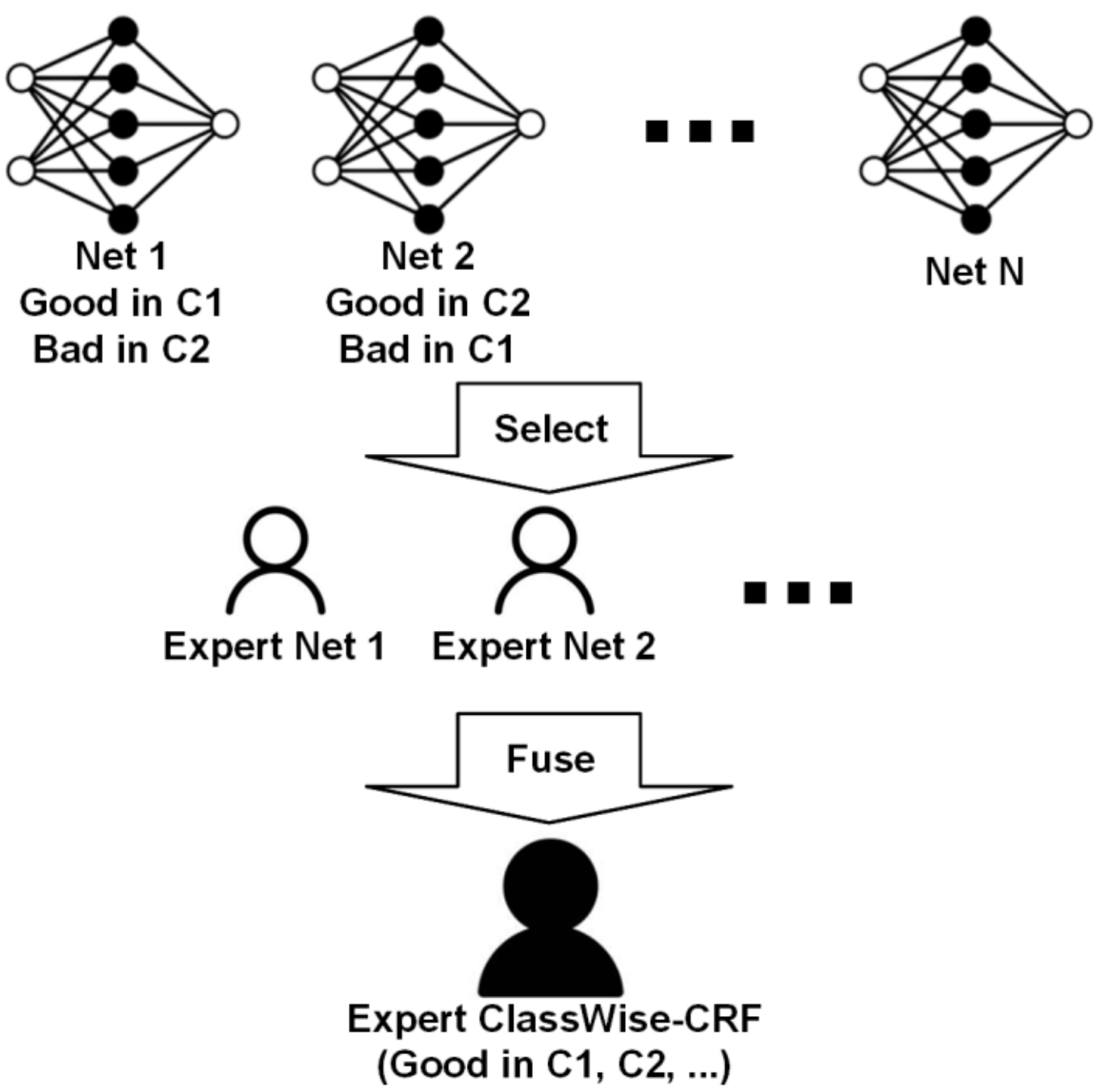

**Fig. 1**: Framework for category-specific expert selection and fusion in semantic segmentation.

Although a single network architecture may exhibit superior overall performance relative to others, it is commonly observed that different architectures display inconsistent strengths across various semantic categories. For example, Net 1 might excel in Category 1 (C1) but underperform in Category 2 (C2), while Net 2 demonstrates the reverse pattern. This inherent variability poses a critical challenge: no individual network can consistently achieve optimal accuracy across all categories in complex tasks like remote sensing image segmentation. Motivated by this limitation, we introduce ClassWise-CRF, a novel result-level fusion framework designed to harness the complementary expertise of multiple networks. By first selecting a subset of high-performing "expert" networks from a diverse candidate pool and then fusing their predictions, ClassWise-CRF aims to deliver balanced, category-optimized segmentation results across all classes, such as C1, C2, C3, and more, as illustrated in **Fig. 1**. Specifically, the framework leverages an efficient greedy algorithm to identify a predefined number of expert networks from the candidates. Drawing inspiration from Conditional Random Fields (CRF) [11, 23], we model the predictions of these experts as confidence vector fields, incorporating segmentation metrics from the validation set as prior weights and applying an exponential weighting strategy for adaptive fusion. Although CRF was originally designed to optimize semantic segmentation boundaries by modeling spatial relationships, we have found it to be the desirable refinement technique for our fusion method, as it naturally integrates with the fused confidence vector fields to enhance reliability, mitigate local distortions from multi-network discrepancies, and ensure smoother category transitions. The resulting fused vector field is then optimized using unary and pairwise potentials in CRF to ensure spatial consistency and boundary accuracy [24] in the final segmentation results. We validate the effectiveness of this architecture using eight semantic segmentation networks on the LoveDA [14] and Vaihingen datasets, with extensive ablation studies confirming the accuracy and robustness of our framework. Theoretically, this approach is not limited to remote sensing image segmentation; it can enhance semantic segmentation performance across various task scenarios, provided inference time constraints are satisfied. As network architectures continue to evolve, the ClassWise-CRF framework, with its plug-and-play characteristics, promises to continually optimize the segmentation performance using state-of-the-art networks.

Our main contributions can be summarized as follows:

1. We propose a general semantic segmentation result-level fusion framework ClassWise-CRF, which adaptively fuses predictions from selected expert networks to achieve optimal segmentation performance across all semantic categories.

2. We introduce a network selection algorithm that adaptively chooses a specified number of expert networks from multiple candidates.
3. We develop a category-specific fusion method, seamlessly integrated with the CRF optimization, to ensure spatial consistency and boundary accuracy in the fused segmentation images.
4. We provide a comprehensive analysis and discussion of the experimental results, along with directions for future research.

The remainder of this paper is organized as follows: Section 2 reviews related work on information fusion strategies and class-wise approaches. Section 3 elaborates on the methodology of ClassWise-CRF. Section 4 describes the experiments. Section 5 analyzes the results, discusses the limitations of the method, and suggests future research directions. Finally, Section 6 concludes the paper.

# 2 Related work

## 2.1 Information fusion in semantic segmentation

Information fusion plays an important role in semantic segmentation, deeply integrated into the evolution of segmentation architectures [25]. By integrating features across multiple dimensions, this technique seeks to enhance the precision and robustness of segmentation models [26, 27]. With the progression of deep learning architectures, strategies such as multi-modal fusion [28], cross-layer feature fusion [29], multi-scale fusion [30, 31], and contextual information fusion [32] have demonstrated substantial performance gains across diverse scenarios.

Multimodal fusion strategies have gained traction for RGB-D or RGB-Thermal urban scene parsing, where complementary modalities like depth or thermal data mitigate limitations in visible-spectrum imaging under adverse conditions. For example, MFFENet [28] and PGDENet [33] employ progressive multi-scale and guided fusion to enhance cross-modal integration, while MTANet [34] and EGFNet [35] use hierarchical and edge-aware mechanisms for better semantic understanding. MDNet [36] introduces Mamba-based diffusion for efficient long-range modeling, and FCDENet [37] leverages feature contrast differences for indoor RGB-D classification. These methods synthesize RGB and supplementary data to improve robustness in low-light or complex environments.

Cross-layer feature fusion forms the backbone of many segmentation networks, combining feature information from different hierarchical levels to achieve a balance between detail retention and semantic expression [29]. This approach enriches the decoder's input by merging high-resolution spatial details from shallow layers with robust semantic representations from deeper layers, yielding more accurate segmentation maps. Architectures like UNet [10, 38] and DeepLabV3+ [11] leverage nested skip connections and multi-level feature fusion networks to dynamically integrate shallow and deep features. These methods preserve boundary details while enhancing adaptability to complex scenes. Similarly, DenseNet [39] and HRNet [40] employ dense connections to facilitate information reuse across layers, further strengthening feature integration.

Multi-scale fusion tackles the challenge of segmenting objects of varying sizes by integrating information from different receptive fields, capturing multi-scale features [31]. This strategy is particularly effective in scenarios like remote sensing image analysis, where object scales differ significantly from small buildings to expansive vegetation coverage. Atrous Spatial Pyramid Pooling (ASPP) [11] employs dilated convolutions with varying rates to extract multi-scale contextual information. Likewise, PSPNet [41] utilizes a pyramid pooling module to perform multi-scale pooling on feature maps, enhancing the model's ability to synthesize global and local information. Although multi-scale fusion methods significantly enhance robustness to scale variations and improve overall segmentation precision, they can sometimes result in redundant computations or artifacts if dilation rates are not carefully tuned, potentially leading to inefficiencies or suboptimal performance in highly dynamic environments.

In recent years, architectures inspired by large language models have been successfully adapted to computer vision tasks, offering new possibilities for contextual information fusion. Models such as Squeeze-and-Excitation (SE) and Convolutional Block Attention Module (CBAM) incorporate channel and spatial attention mechanisms, significantly boosting semantic segmentation performance. Additionally, ViTs and Vision Mamba [21, 42] leverage long-range dependencies to integrate global context, improving category recognition accuracy [43]. Another emerging trend is architecture fusion, which combines the strengths of distinct network designs. For instance, fusions of CNNs and ViTs capitalize on CNNs' local feature extraction capabilities and ViTs' global modeling prowess, achieving breakthroughs in high-resolution image segmentation tasks [44, 45]. Similarly, the VMamba[21] architecture integrates the Mamba framework with CNNs to optimize performance. Recent works like LSENet [46] enhance the Swin Transformer with spatial enhancement modules (SEM) and local enhancement modules (LEM) to improve feature extraction and edge accuracy in remote sensing semantic segmentation, achieving strong mIoU scores on datasets such as Vaihingen and Potsdam. Likewise, the Swin Transformer with Spatial and Local Context Augmentation [47] builds on this by incorporating similar enhancements, further extending performance to additional datasets like OpenEarthMap. These advanced fusion and attention-based approaches provide superior handling of global contexts and hybrid features, leading to state-of-the-art results in challenging tasks. However, they frequently entail higher model complexity, longer training times, and greater parameter counts. In addition, existing fusion strategies predominantly focus on feature-level integration [26], often neglecting the category-specific differences inherent in semantic segmentation. In practice, different networks exhibit varying segmentation performance across specific semantic categories [48], a subtle variation that feature-level fusion alone may not fully address.

In addition to architecture fusion, a few studies have explored model ensembling, which bears resemblance to our work in leveraging multiple networks for improved segmentation. For instance, Dimitrovski et al. [49] investigated the use of three backbone networks within the U-Net architecture, fusing their segmentation outputs via a geometric mean ensemble method to generate the final predictions. However, this approach lacks category-specific processing, failing to address class-level performance variations and relying solely on uniform averaging. Similarly, Li et al. [50] applied model ensembling in a remote sensing image copyright protection framework, extracting robust semantic features from two segmentation models and fusing them through a voting mechanism, followed by hash-based similarity detection to identify tampering or infringement. Yet, this method employs simple ensemble learning (i.e., voting) without category-specific weights or expert selection. In contrast, our proposed ClassWise-CRF offers greater flexibility through a candidate pool and enables more reliable fusion by adaptively weighting contributions based on per-category performance, thereby optimizing results across diverse semantic classes.

## 2.2 Class-wise strategies in semantic segmentation

Class-wise strategies have emerged in semantic segmentation to address category-specific challenges by tailoring processing or fusion techniques to individual classes [51], thereby enhancing overall segmentation performance [52]. For instance, Class-Wise Fully Convolutional Network (C-FCN) [53] is an architecture that emphasizes independent decoding and classification for each category. Specifically, the network employs a shared encoder to extract general features, followed by category-specific decoders and classifiers for per-class processing. This approach has shown promising results in remote sensing semantic segmentation. Other strategies model intra-class relationships using graph structures. For instance, Class-Wise Dynamic Graph Convolution (CDGC) [54] introduces a module that constructs graphs among pixels of the same category, applying dynamic graph convolutions to aggregate features within each class. This method reduces inter-class interference, thereby improving segmentation accuracy. Although CDGC excels in minimizing inter-class noise and boosting intra-class coherence, its graph-based operations can be computationally intensive, potentially slowing down inference in large-scale applications. Similarly, EDC$^3$ [55] presents an ensemble approach based on class-specific Copula functions, modeling statistical dependencies between classifiers to achieve class-wise fusion and enhance performance. However, these methods often significantly increase model complexity, impose high computational resource demands, and exhibit poor scalability, which limits their practical application in large-scale or resource-constrained environments.

In contrast to these methods, our proposed ClassWise-CRF explores result-level category-specific fusion. By adaptively selecting and integrating predictions from networks that excel in specific categories, and further refining the results through the CRF optimization, our approach ensures spatial consistency and boundary accuracy. ClassWise-CRF offers a simple yet effective means to boost performance in semantic segmentation, addressing the limitations of feature-level fusion and existing class-wise strategies. Additionally, the architecture exhibits strong scalability due to its modular design and independence from specific network architectures. It paves the way for new research directions in semantic segmentation by providing a framework that complements and extends current methodologies.

# 3 ClassWise-CRF

## 3.1 Network selection

For semantic segmentation of remote sensing images, a variety of architectures are currently available, each demonstrating distinct strengths and weaknesses in their performance across different semantic categories. However, directly fusing the predictions from all networks is impractical due to the substantial computational overhead and the potential risks of introducing redundancy or noise. Consequently, the ClassWise-CRF architecture proposes the selection of a small subset of suitable networks for fusion to enhance overall segmentation performance. This section presents a greedy-algorithm-based strategy for network selection, with the objective of selecting a subset ($K$ networks) from the total pool ($N$ networks) to maximize the fused mean Intersection over Union (mIoU) score. The selection strategy rests on the assumption that the fused result for each category should, as closely as possible, match the Intersection over Union (IoU) score of the best-performing network for that category. To achieve this, we have developed an efficient greedy algorithm for progressively network selection, as outlined in **Algorithm 1**.

**Algorithm 1** Network Selection Algorithm

**Input:** $iou_matrix$: $N \times c$ IoU matrix, $K$: number of networks to select, $methods$: list of network names

**Output:** List of selected network names

```
selected_indices ← ∅
for i = 1 to K do
    best_miou ← −1
    best_net ← −1
    for each j in 0 to N − 1 where j ∉ selected_indices do
        current_selected ← selected_indices ∪ {j}
        max_iou_per_class ← max(iou_matrix[current_selected], axis = 0)
        miou ← mean(max_iou_per_class)
        if miou > best_miou then
            best_miou ← miou
            best_net ← j
        end if
    end for
    selected_indices ← selected_indices ∪ {best_net}
end for
return [methods[i] for i in selected_indices]
```

Specifically, given an IoU score matrix iou_matrix $\in R^{n \times c}$, where $n$ represents the number of networks and $c$ denotes the number of categories, with iou_matrix$[i, j]$ indicating the IoU score of the $i$-th network on the $j$-th category, the algorithm aims to iteratively select $K$ networks to maximize the fused mIoU. The value of hyperparameter $K$ can be determined according to the requirements of a specific scenario or user preference. In general, when candidate networks show varying strong performance in different categories, increasing the value of $K$ is likely to result in better fusion results in subsequent segmentation. The algorithm initializes with an empty set and iterates $K$ times. In each iteration, it traverses all unselected networks, computes the mIoU resulting from adding each candidate network to the current set of selected networks, and chooses the one that provides the greatest improvement in mIoU. For each category, the fused IoU score is determined as the highest IoU score achieved by the selected networks for that category, and the mIoU is calculated as the average of these maximum IoU scores across all categories. The algorithm ultimately returns the name list of the $K$ selected networks.

Our proposed network selection algorithm is an efficient greedy-based approach, despite the resulting selection may not always be optimal. It offers the advantage of enhanced computational efficiency with a near-optimal selection, making it particularly suitable when the pool of candidate networks is large, as discussed in Section 4.4.5. However, when the pool of candidate networks is relatively small, one can consider using an exhaustive search to achieve an optimal network selection, as the difference in time consumption between the two methods would be insignificant. Since the network selection algorithm is performed during the deployment phase and not during real-time inference, the inference efficiency of our proposed ClassWise-CRF is high.

## 3.2 Category-specific fusion

After selecting the desired $K$ networks, the subsequent step involves integrating the prediction results of these $K$ networks. To this end, we first obtain the IoU scores for each of the selected $K$ networks on the validation set, computed individually for every semantic category. These IoU scores serve as prior information, quantifying the performance of each network on specific categories. The core idea of category-specific fusion is to dynamically adjust the contribution weight of each network based on its IoU performance for a given category. Specifically, in the fusion process, a network with a higher IoU score in a particular category is assigned a greater weight for that category. This approach ensures that the network with superior performance dominates the segmentation of its corresponding category in our proposed framework, thereby achieving complementary and optimized segmentation results across the different networks.

Each network's prediction result for an input image is represented as a probability map, which provides the confidence distribution across all categories for each pixel, rather than a binarized label. The purpose is to preserve the complete uncertainty information of the network's output, as directly using binarized labels may lead to significant bias due to information loss. Suppose there are $K$ networks and a total of $C$ categories. Each network $k$ (where $k = 1, 2, \dots, K$) produces a probability prediction map $P_k \in R^{C \times H \times W}$, where $P_{k,c}(i,j)$ denotes the confidence of network $k$ for category $c$ at pixel $(i,j)$, satisfying $\sum_{c=1}^{C} P_{k,c}(i,j) = 1$. The objective of this strategy is to fuse these probability maps to generate a unified probability map $P \in R^{C \times H \times W}$ that maximizes segmentation performance.

Assuming the IoU score of network $k$ for category $c$ is $\mathrm{IoU}_{k,c}$, to amplify performance differences, we adopt an exponential weighting strategy by introducing a hyperparameter $\alpha > 1$ to control the degree of weight amplification. The weight $w_{k,c}$ of network $n$ for category $c$ is defined as:

$$w_{k,c} = \frac{\mathrm{IoU}_{k,c}^{\alpha}}{\sum_{m=1}^{K} IoU_{m,c}^{\alpha}} \tag{1}$$

Typically, α is set within the range [1.0, 3.0], determined through experimentation. When $\alpha = 1$, the weighting reduces to linear weighting; when $\alpha > 1$, networks with higher IoU scores receive significantly larger weights for that category, thereby enhancing their influence on the fusion result. The weights satisfy the normalization condition:

$$\sum_{k=1}^{K} w_{k,c} = 1, \qquad \forall c \in \{1,2,\dots,C\} \tag{2}$$

Based on the aforementioned weights, the confidence $P_c(i,j)$ for category $c$ at pixel $(i,j)$ in the fused probability map $P$ is calculated as:

$$P_c(i,j) = \sum_{k=1}^{K} w_{k,c} \cdot P_{k,c}(i,j) \tag{3}$$

To ensure that the fusion result satisfies the probability distribution property (i.e., $\sum_{c=1}^{C} P_c(i,j) = 1$), the confidence for each pixel is normalized:

$$P_c(i,j) = \frac{\sum_{k=1}^{K} w_{k,c} \cdot P_{k,c}(i,j)}{\sum_{m=1}^{C} \sum_{k=1}^{K} w_{k,m} \cdot P_{k,m}(i,j)} \tag{4}$$

Thus, we obtain the fused probability map. However, due to potential significant discrepancies in predictions from different networks, direct fusion may lead to local distortions or inconsistencies. Therefore, further optimization introduced in Section 3.3 is necessary to enhance the accuracy and spatial consistency of the segmentation results.

## 3.3 CRF optimization

In the process of category-specific fusion, although we generate a high-quality probability map $P$ through multi-network collaboration, the spatial distribution of land cover categories in remote sensing images often exhibits strong correlations. The fused probability map $P_c(i,j)$ may not fully capture these relationships, particularly in regions with complex textures or category transitions. Due to intricate spatial structures and subtle differences in predictions from different networks, the fusion results may exhibit noise or inconsistencies in local areas or at category boundaries. To address this issue and further enhance the spatial continuity and boundary precision of the segmentation results, we draw inspiration from CRF [23, 56]. We find that the CRF optimization strategy aligns well with our fused probability map. By integrating CRF with the probability map, we leverage its ability to model contextual relationships between pixels, thereby improving the overall consistency and accuracy of the segmentation map [11, 57].

In the ClassWise-CRF architecture, CRF serves as a critical optimization step following category-specific fusion. Specifically, the CRF optimization process, as depicted in Algorithm 2, constructs unary and pairwise terms using the initial probability map $P_c(i,j)$ and image features. These terms are used to formulate an energy function, which is iteratively minimized to progressively enhance the spatial continuity of a segmented image while maintaining its consistency with the original remote sensing image.

**Algorithm 2** CRF Optimization Algorithm

**Input:** Fusion probability map $\mathbf{P} \in \mathbb{R}^{C\times H\times W}$, image $\mathbf{I}$, CRF parameters $(\sigma_g, \sigma_b, \sigma_c, w_g, w_b)$

**Output:** Optimized label map $\mathbf{x}^*$

Initialize $\mathbf{x}$ by $\arg\max_c P_c(i,j)$ for each pixel $(i,j)$

Compute unary potentials: $\psi_u(x_{i,j} = c) = -\log(P_c(i,j))$ for all $c, i, j$

Calculate pairwise potentials:

$$k_g(\mathbf{f}_{i,j}, \mathbf{f}_{k,l}) = \exp\left(-\frac{|(i,j)-(k,l)|^2}{2\sigma_g^2}\right)$$

$$k_b(\mathbf{f}_{i,j}, \mathbf{f}_{k,l}) = \exp\left(-\frac{|(i,j)-(k,l)|^2}{2\sigma_b^2} - \frac{|\mathbf{I}_{i,j}-\mathbf{I}_{k,l}|^2}{2\sigma_c^2}\right)$$

$$\psi_p(x_{i,j}, x_{k,l}) = [x_{i,j} \neq x_{k,l}]\left[w_g k_g(\mathbf{f}_{i,j}, \mathbf{f}_{k,l}) + w_b k_b(\mathbf{f}_{i,j}, \mathbf{f}_{k,l})\right]$$

Formulate energy function: $E(\mathbf{x}) = \sum_{i,j} \psi_u(x_{i,j}) + \sum_{(i,j)<(k,l)} \psi_p(x_{i,j}, x_{k,l})$

Minimize $E(\mathbf{x})$ using mean-field approximation to obtain $\mathbf{x}^*$

**return** $\mathbf{x}^*$

The unary potential $\psi_u(x_{i,j})$ directly reflects the probability that each pixel $(i,j)$ independently belongs to a certain category $c$, and can be expressed as:

$$\psi_u(x_{i,j} = c) = -\log(P_c(i,j)) \quad (5)$$

where $x_{i,j}$ denotes the category label of pixel $(i,j)$, and $P_c(i,j)$ is the confidence of the fused probability map for category $c$ at that pixel.

The pairwise potential $\psi_p(x_{i,j}, x_{k,l})$ captures the relationships between neighboring pixels, encouraging pixels that are spatially and colorimetrically similar to share the same category label. We employ a combination of Gaussian and bilateral kernels for this purpose. The Gaussian kernel, based on the spatial distance between pixels, aims to enforce spatial continuity by promoting identical labels for pixels in close proximity. It is defined as:

$$k_g(\boldsymbol{f}_{i,j}, \boldsymbol{f}_{k,l}) = \exp\left(-\frac{|(i,j)-(k,l)|^2}{2\sigma_g^2}\right) \quad (6)$$

where $(i,j)$ and $(k,l)$ represent the spatial coordinates of the pixels, and $\sigma_g$ controls the range of spatial correlation.

The bilateral kernel incorporates both spatial distance and color differences, aiming to achieve edge-aware smoothing. It encourages smoothing in regions with similar colors while preserving boundary information. The bilateral kernel is defined as:

$$k_b(f_{i,j}, f_{k,l}) = \exp\left(-\frac{|(i,j)-(k,l)|^2}{2\sigma_b^2} - \frac{|I_{i,j} - I_{k,l}|^2}{2\sigma_c^2}\right) \quad (7)$$

where $I_{i,j}$ and $I_{k,l}$ are the color values of pixels $(i,j)$ and $(k,l)$, respectively, and $\sigma_b$ and $\sigma_c$ control the influence of spatial and color differences. Thus, the pairwise potential can be defined as:

$$\psi_p(x_{i,j}, x_{k,l}) = \mu(x_{i,j}, x_{k,l})[w_g k_g(\boldsymbol{f}_{i,j}, \boldsymbol{f}_{k,l}) + w_b k_b(\boldsymbol{f}_{i,j}, \boldsymbol{f}_{k,l})] \quad (8)$$

where $\mu(x_{i,j}, x_{k,l}) = [x_{i,j} \neq x_{k,l}]$ is the Potts model [58], which penalizes neighboring pixels with different labels, and $w_g$ and $w_b$ are the weights for the Gaussian and bilateral kernels, respectively.

The energy function $E(x)$ of the CRF combines the unary and pairwise potentials, defined as:

$$E(x) = \sum_{i,j} \psi_u(x_{i,j}) + \sum_{(i,j)<(k,l)} \psi_p(x_{i,j}, x_{k,l}) \tag{9}$$

Substituting $P_c(i,j)$, the energy function takes the specific form:

$$E(x) = \sum_{i,j} -\log\left(P_{x_{i,j}}(i,j)\right) + \sum_{(i,j)<(k,l)} \mu(x_{i,j}, x_{k,l})\left[w_g k_g(\boldsymbol{f}_{i,j}, \boldsymbol{f}_{k,l}) + w_b k_b(\boldsymbol{f}_{i,j}, \boldsymbol{f}_{k,l})\right] \tag{10}$$

The optimization objective is to find the label configuration $x^*$ that minimizes $E(x)$ using the mean-field approximation algorithm:

$$x^* = \arg\min_x E\,(x) \tag{11}$$

To ensure the applicability of CRF, we employ the Bayesian optimization to automatically tune key parameters, thereby achieving dynamic modeling of spatial and color relationships.

# 4 Experiments

## 4.1 Datasets and metrics

To validate the effectiveness of ClassWise-CRF, we conducted experiments on two representative remote sensing semantic segmentation datasets: LoveDA [14] and Vaihingen. The LoveDA dataset comprises 5987 high spatial resolution (0.3 m) remote sensing images, covering the cities of Nanjing, Changzhou, and Wuhan. It focuses on the impact of urban-rural geographical environmental differences on semantic segmentation and domain adaptation tasks. This dataset presents three major challenges: multi-scale objects, complex background samples, and inconsistent class distributions. It includes 2522 training images, 1669 validation images, and 1796 test images, all with a resolution of 1024×1024 pixels. The dataset is annotated with seven categories: background, building, road, water, barren, forest, and agricultural. In our experiments, prior results from the validation set were used for parameter optimization, while the test set was used for inference, with performance evaluated through the official online evaluation system.

The Vaihingen dataset consists of 33 remote sensing images with a spatial resolution of 9 cm, each averaging 2494×2064 pixels in size, and includes near-infrared, red, and green bands. It is annotated with six categories: impervious surface, building, low vegetation, tree, car, and clutter. In our experiments, the dataset was split as follows: training images with IDs 1, 3, 5, 7, 11, 13, 15, 17, 21, 23, 26, 28, 30, 32, 34, and 37; validation images with IDs 2, 4, 6, 8, 10, 12, 14, 16, and 20; and test images with IDs 22, 24, 27, 29, 31, 33, 35, and 38. When calculating mIoU, the clutter category was excluded. Consistent with the LoveDA dataset, prior results from the validation set were used for parameter optimization, while the test set was used for final inference.

We employ IoU and mIoU as evaluation metrics. IoU measures the segmentation accuracy for each category and is calculated using the formula:

$$\mathrm{IoU}_c = \frac{\mathrm{TP}_c}{\mathrm{TP}_c + \mathrm{FP}_c + \mathrm{FN}_c} \tag{12}$$

Where $\mathrm{TP}_c$, $\mathrm{FP}_c$, and $\mathrm{FN}_c$ represent the number of true positive, false positive, and false negative pixels for category $c$, respectively. The mIoU is the average IoU across all categories, computed as:

$$\mathrm{mIoU} = \frac{1}{C}\sum_{c=1}^{C} IoU_c \tag{13}$$

where $C$ denotes the total number of categories.

## 4.2 Results

In this study, we considered eight representative semantic segmentation architectures for experimentation, including Segformer [59], combinations of UperNet [60] with ResNet [61], PSPNet [41] with ResNet, DeepLabV3+ [11] with ResNet, BiseNet [62] with ResNet, UperNet with ConvNeXt [20], UperNet with VMamba [21], and UperNet with Swin [19]. The semantic segmentation performance of these architectures on the validation sets of the LoveDA and Vaihingen datasets is detailed in **Table 1** and **Table 2**, respectively. By applying Algorithm 1 (Network Selection) with the parameter $K = 3$, we identified ConvNeXt, VMamba, and Swin as the networks for subsequent fusion in both the LoveDA and Vaihingen datasets. Hereafter, ConvNeXt, VMamba, and Swin refer to their combinations with UperNet for semantic segmentation. In the LoveDA dataset, ConvNeXt performed best on the background, building, and agricultural classes; VMamba exceled in the road, water, and barren classes; and Swin was optimal for the forest class. In the Vaihingen dataset, ConvNeXt was superior for the car class, VMamba for the vegetation and tree classes, and Swin for the surface and building classes. Amongst these three networks, each semantic category was best handled by one of them. This observation supported choosing a value of 3 for $K$ and further validated the effectiveness of Algorithm 1 in identifying the appropriate expert networks.

**Table 1.** Segmentation performance of candidate networks in the LoveDA validation set. The highest score in the IoU metric is shown in bold.

| Decoder | Encoder | background | building | road | water | barren | forest | agricultural | mIoU |
|---|---|---|---|---|---|---|---|---|---|
| Segformer-b0 | MixViT | 53.23 | 62.71 | 52.12 | 63.40 | 29.37 | 41.93 | 45.65 | 49.77 |
| Upernet | ResNet50 | 54.04 | 60.14 | 50.21 | 59.86 | 25.00 | 40.84 | 50.24 | 48.62 |
| PSPNet | ResNet50 | 53.85 | 60.57 | 51.37 | 58.76 | 21.92 | 42.31 | 50.41 | 48.46 |
| DeepLabV3+ | ResNet50 | 53.44 | 61.63 | 53.50 | 62.30 | 26.02 | 42.69 | 49.26 | 49.83 |
| BiSeNetv1 | ResNet50 | 53.31 | 55.97 | 50.58 | 59.85 | 27.75 | 40.15 | 46.74 | 47.76 |
| Upernet | ConvNeXt-t | **55.58** | **66.08** | 56.32 | 68.61 | 32.82 | 42.33 | **55.88** | **53.95** |
| Upernet | VMamba-t | 55.29 | 65.08 | **56.89** | **71.31** | **33.68** | 40.17 | 53.52 | 53.71 |
| Upernet | Swin-t | 54.61 | 65.37 | 55.98 | 69.79 | 31.69 | **44.14** | 52.19 | 53.40 |

**Table 2.** Segmentation performance of candidate networks in the Vaihingen validation set. The highest score is shown in bold.

| Decoder | Encoder | surface | building | vegetation | tree | car | clutter | mIoU |
|---|---|---|---|---|---|---|---|---|
| Segformer-b0 | MixViT | 81.46 | 88.24 | 66.25 | 81.18 | 62.51 | 50.72 | 75.93 |
| Upernet | ResNet50 | 80.86 | 88.59 | 64.31 | 80.05 | 52.63 | 50.26 | 73.29 |
| PSPNet | ResNet50 | 81.01 | 88.00 | 63.81 | 79.40 | 48.46 | 50.01 | 72.14 |
| DeepLabV3+ | ResNet50 | 80.19 | 87.53 | 63.23 | 79.35 | 51.35 | 50.70 | 72.33 |
| BiSeNetv1 | ResNet50 | 81.50 | 89.04 | 65.99 | 80.41 | 55.65 | **50.75** | 74.52 |
| Upernet | ConvNeXt-t | 83.35 | 90.65 | 68.75 | 82.63 | **72.57** | 49.68 | 79.59 |
| Upernet | VMamba-t | 83.39 | 91.11 | **68.91** | **83.04** | 70.46 | 49.51 | 79.38 |
| Upernet | Swin-t | **84.00** | **91.16** | 68.55 | 82.67 | 72.03 | 50.36 | **79.68** |

Next, parameter optimization was conducted on the validation set, and the segmentation performance after Bayesian optimization is presented in **Table 3** and **Table 4**. The value of α is set as 2.5, as discussed in Section 4.3.2. It can be observed that when $K = 3$, our framework outperformed the best baseline by 1.00% in mIoU on the LoveDA dataset and by 0.87% on the Vaihingen dataset during validation set optimization. Moreover, for most categories, the IoU scores of the fused results exceeded those of all baseline methods. This consistent mIoU gain underscores the framework's ability to leverage complementary strengths across networks, achieving a more balanced performance profile that is particularly valuable in remote sensing applications where category imbalances are common. For a more comprehensive experiment, we have also conducted experiments on pairwise combinations in our study. A similar trend was observed in pairwise combinations, where the mIoU scores consistently surpassed those of the corresponding baseline methods. It can be intuitively observed that the overall semantic segmentation performance when $K = 3$ is superior to the pairwise combinations.

Although integrating more networks (e.g., $K = 3$ ) enhanced the overall performance in terms of mIoU, the fused results were inevitably influenced by more networks, including those that may underperform on certain categories. Consequently, fusing three networks ($K = 3$) might yield inferior performance in certain individual categories compared to a pairwise network combination. For example, the fusion of ConvNeXt and VMamba achieved higher IoU scores for the "road" class in Table 3 and the "car" class in Table 4, compared to the fusion of the three networks. In such cases, the inclusion of an additional network (e.g., Swin) introduced predictions that diluted the fused confidence for categories where it underperformed, leading to slight IoU drops (e.g., 57.91% for "road" in C+V vs. 57.84% in C+V+S on LoveDA validation). Similarly, for categories like "forest" and "agricultural" in Table 3, the fused IoU (42.62% and 54.58%, respectively) may fall short of the best pairwise combinations due to inconsistent expert performance; for instance, VMamba's IoU on "forest" was 3.97% lower than Swin's (40.17% vs. 44.14%), thereby exerting a negative influence on the exponential-weighted fusion. Additionally, we have compared our fusion results against well-known semantic segmentation models such as ABCNet and TransUNet as additional baselines, with the fused performance (e.g., C+V+S) outperforming these models in mIoU and most per-category IoUs. The segmentation results for ABCNet [63], and TransUNet [64] were as reported in [22]. These instances highlight a key limitation: while the method prioritized the global mIoU optimization, it might trade off marginal per-category gains in scenarios with high variance among expert networks. In applications, this is not a significant concern, as our primary goal is to elevate the overall performance of a network (or fused networks in our study). Nevertheless, for tasks emphasizing minority or challenging categories, users could adjust $K$ to smaller values or refine the selection algorithm to prioritize variance minimization.

**Table 3.** Comparison of the segmentation performance of ClassWise-CRF and baseline methods on the LoveDA validation set. The highest score is shown in bold. Hereafter, “C” represents “ConvNeXt”; “V” represents “VMamba”; “S” represents “Swin”.

| Strategy | Methods | background | building | road | water | barren | forest | agricultural | mIoU |
|---|---|---|---|---|---|---|---|---|---|
| Baseline | C | 55.58 | 66.08 | 56.32 | 68.61 | 32.82 | 42.33 | **55.88** | 53.95 |
| | V | 55.29 | 65.08 | 56.89 | 71.31 | 33.68 | 40.17 | 53.52 | 53.71 |
| | S | 54.61 | 65.37 | 55.98 | 69.79 | 31.69 | **44.14** | 52.19 | 53.40 |
| | ABCNet | 35.15 | 46.37 | 48.61 | 44.31 | 28.43 | 37.58 | 11.10 | 40.58 |
| | TransUNet | 35.13 | 58.90 | 57.57 | 67.61 | 25.93 | 37.01 | 20.85 | 49.23 |
| Pairwise Combinations | C+V | **56.25** | 67.06 | **57.91** | 72.02 | **34.17** | 41.39 | 55.42 | 54.89 |
| | C+S | 55.94 | 67.53 | 57.24 | 70.91 | 32.99 | 43.71 | 54.84 | 54.74 |
| | S+V | 55.61 | 66.78 | 57.81 | 71.68 | 33.80 | 41.94 | 53.53 | 54.45 |
| $K = 3$ | C+V+S | 56.14 | **67.64** | 57.84 | **72.11** | 33.76 | 42.62 | 54.58 | **54.96** |

**Table 4.** Comparison of the segmentation performance of ClassWise-CRF and baseline methods on the Vaihingen validation set. The highest score is shown in bold.

| Strategy | Methods | surface | building | vegetation | tree | car | clutter | mIoU |
|---|---|---|---|---|---|---|---|---|
| Baseline | C | 83.35 | 90.65 | 68.75 | 82.63 | 72.57 | 49.68 | 79.59 |
| | V | 83.39 | 91.11 | 68.91 | 83.04 | 70.46 | 49.51 | 79.38 |
| | S | 84.00 | 91.16 | 68.55 | 82.67 | 72.03 | 50.36 | 79.68 |
| | ABCNet | 81.45 | 89.21 | 64.59 | 81.95 | 58.80 | 48.21 | 75.20 |
| | TranUNet | 83.10 | 89.25 | 65.32 | 82.70 | 70.45 | 49.40 | 78.16 |
| Pairwise Combinations | C+V | 84.29 | 91.18 | 69.94 | 83.46 | **73.83** | 50.08 | 80.54 |
| | C+S | 84.41 | 91.42 | 69.87 | 83.34 | 73.27 | 50.06 | 80.46 |
| | S+V | 84.36 | 91.48 | 69.89 | 83.42 | 73.35 | **50.38** | 80.50 |
| $K = 3$ | C+V+S | **84.53** | **91.53** | **70.23** | **83.53** | 72.93 | 50.24 | **80.55** |

After fixing the hyperparameters on the validation set, we applied the ClassWise-CRF framework to the test set to evaluate its generalization ability. The experimental results on the test set are listed in **Table 5** and **Table 6**. The findings indicate that the framework achieved performance levels comparable to or exceeding the best baseline methods across all categories, and outperformed all baselines in most categories. Specifically, when $K = 3$, the mIoU score on the LoveDA dataset improved by 0.68% compared to the best baseline, and by 0.91% on the Vaihingen dataset, fully demonstrating the superiority and robustness of the ClassWise-CRF framework in practical applications. Notably, the "barren" category in LoveDA shows modest IoU gains (e.g., from 32.82% to 33.76% on validation, and 15.96% to 16.92% on test), which were smaller than those in other categories like "building" (from 66.08% to 67.64%). This can be attributed to the inherent challenges of minority categories in remote sensing data, such as "barren", which often exhibited sparse representations, high intra-class variability (e.g., diverse textures from arid lands), and class imbalance in datasets like LoveDA. Our fusion method adapted well to such categories by prioritizing the best-performing expert (e.g., VMamba at 33.68% on validation), but the exponential weighting may amplify subtle discrepancies among experts, limiting gains when baselines are already low and variable.

**Table 5.** Comparison of the segmentation performance of ClassWise-CRF and baseline methods on the LoveDA test set. The highest score is shown in bold.

| Strategy | Methods | background | building | road | water | barren | forest | agricultural | mIoU |
|---|---|---|---|---|---|---|---|---|---|
| Baseline | C | 44.47 | 57.74 | 55.05 | 78.28 | 15.96 | 45.61 | 61.49 | 51.23 |
| | V | 45.55 | 58.81 | 58.79 | 80.39 | 16.99 | 46.99 | 62.95 | 52.92 |
| | S | 46.33 | 56.96 | 56.63 | 79.54 | 16.88 | 46.20 | 62.32 | 52.12 |
| Pairwise Combinations | C+V | 46.82 | 59.31 | 59.36 | 80.44 | 16.34 | 47.14 | 63.41 | 53.26 |
| | C+S | 47.10 | 58.50 | 57.99 | 79.76 | 16.68 | 46.81 | 62.88 | 52.82 |
| | S+V | 47.48 | 58.98 | **59.53** | **80.82** | **16.99** | 47.39 | 63.73 | 53.56 |
| $K = 3$ | C+V+S | **47.70** | **59.35** | 59.48 | 80.57 | 16.92 | **47.41** | **63.75** | **53.60** |

**Table 6.** Comparison of the segmentation performance of ClassWise-CRF and baseline methods on the Vaihingen test set. The highest score is shown in bold.

| Strategy | Methods | surface | building | vegetation | tree | car | clutter | mIoU |
|---|---|---|---|---|---|---|---|---|
| Baseline | C | 88.02 | 91.06 | 74.75 | 78.30 | 69.34 | 23.47 | 80.29 |
| | V | 87.99 | 90.34 | 73.57 | 78.27 | 66.29 | **28.72** | 79.29 |
| | S | 88.02 | 90.81 | 74.84 | 78.03 | 68.15 | 27.11 | 79.97 |
| Pairwise Combinations | C+V | 88.41 | 91.75 | 75.97 | 79.17 | **70.25** | 24.02 | 81.11 |
| | C+S | 88.44 | 91.60 | 76.04 | 78.96 | 70.10 | 24.86 | 81.03 |
| | S+V | 88.37 | 91.35 | 75.67 | 78.80 | 69.67 | 27.30 | 80.77 |
| $K = 3$ | C+V+S | **88.58** | **91.91** | **76.37** | **79.21** | 69.95 | 25.51 | **81.20** |

## 4.3 Visual analysis

To intuitively evaluate the performance of the ClassWise-CRF architecture, we compared its segmentation predictions on the LoveDA test set with those of three baseline networks. The specific visualization results are presented in **Fig 2**. Although the ground truth for the LoveDA test set was not directly accessible for quantitative analysis, visual inspection clearly reveals that ClassWise-CRF significantly enhanced segmentation quality by fusing and optimizing the predictions from the three baseline networks. Particularly in regions where the baseline networks produce conflicting predictions, as highlighted by red boxes for emphasis, the method demonstrated an adaptive adjustment capability, steering the final prediction towards the network that excelled in the specific category and yielding more reasonable outcomes without scattered or fragmented artifacts, better aligning with realistic scene structures. This adaptability not only improves inter-category prediction consistency but also bolsters the robustness of the results. Moreover, the prediction maps generated by ClassWise-CRF exhibit excellent edge smoothness, effectively mitigating discrete or jagged errors, thereby enhancing both the visual quality and spatial coherence of the segmentation outcomes.

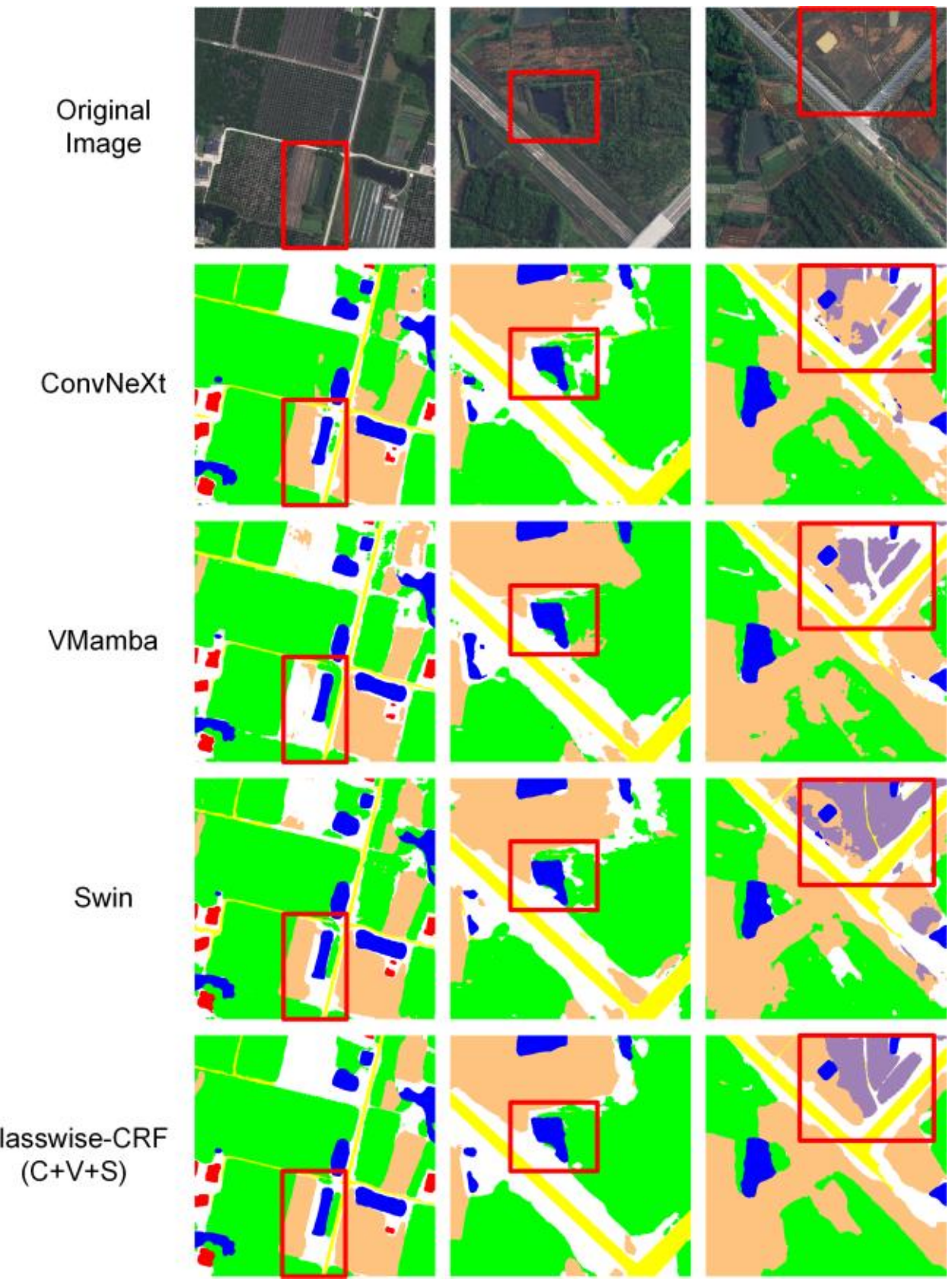

**Fig 2.** Visualization of the segmentation performance of ClassWise-CRF and the baseline methods on the LoveDA test set

## 4.4 Ablation experiments

All subsequent ablation studies in this section were conducted using pairwise combinations of the top three performing baseline networks: ConvNeXt-t, VMamba-t, and Swin-t.

### 4.4.1 Validity of probability plots

In semantic segmentation tasks, the final output of a network typically takes the form of a binarized prediction map, where each pixel is assigned to a single category label. While this binarized form is intuitive, it often introduces significant bias during the fusion process, particularly when weighted by prior segmentation performance metrics (e.g., IoU scores). Specifically, binarized fusion overlooks the network's potential confidence in other categories and heavily relies on the predictions of networks with larger weights, leading to severe category assignment bias. In contrast, probability maps, representing the confidence distribution of each pixel across all categories, preserve the complete uncertainty information of the network's predictions, providing a more robust input for the fusion process.

To fairly compare the effects of binarized outputs and probability maps in fusion, we designed a simple experiment free from prior information interference. This involved directly fusing the prediction results of multiple networks without using segmentation performance priors from the validation set (e.g., IoU scores) and applying CRF optimization. The results on the LoveDA validation set are shown in **Table 7**. The findings indicate that probability map fusion significantly outperforms binarized fusion in terms of the mIoU metric.

**Table 7**. Comparison of segmentation performance on the LoveDA validation set using binarization and probability map strategies. The highest score is shown in bold.

| Strategy | Methods | background | building | road | water | barren | forest | agricultural | mIoU |
|---|---|---|---|---|---|---|---|---|---|
| Baseline | C | 55.58 | 66.08 | 56.32 | 68.61 | 32.82 | 42.33 | **55.88** | 53.95 |
| | V | 55.29 | 65.08 | 56.89 | 71.31 | 33.68 | 40.17 | 53.52 | 53.71 |
| | S | 54.61 | 65.37 | 55.98 | 69.79 | 31.69 | **44.14** | 52.19 | 53.40 |
| Binarization | C+V | 56.00 | 66.71 | 57.79 | **72.27** | 33.41 | 40.75 | 54.85 | 54.54 |
| | C+S | 55.77 | 67.03 | 57.15 | 71.27 | 32.25 | 43.34 | 54.34 | 54.45 |
| | S+V | 55.42 | 66.39 | 57.69 | 71.97 | 32.64 | 41.36 | 53.23 | 54.10 |
| Probability | C+V | **56.20** | 67.00 | 57.80 | 72.04 | **34.14** | 41.26 | 55.20 | **54.81** |
| | C+S | 55.77 | **67.26** | 57.15 | 71.07 | 33.02 | 43.60 | 54.56 | 54.63 |
| | S+V | 55.68 | 66.78 | **57.86** | 71.80 | 33.49 | 41.76 | 53.57 | 54.42 |

## 4.4.2 Influence of α

As elaborated in Section 3.2, our ClassWise-CRF architecture introduces a hyperparameter α to control the degree of weight amplification. To systematically evaluate the impact of α, we conducted extensive experiments on the LoveDA and Vaihingen datasets, with α values ranging from 1.0 to 3.0. The experimental results are detailed in **Table 8** and **Table 9**, respectively. Through data analysis and visualization, we plotted line charts, as shown in the **Fig. 3**, to intuitively demonstrate the effect of varying α on the mIoU metric. The results reveal that within the selected range, all values of α yield performance significantly superior to the baseline methods. Furthermore, the ClassWise-CRF architecture exhibits low sensitivity to the choice of α, with the optimal overall performance achieved when $\alpha = 2.5$.

**Table 8.** Comparison of segmentation performance on the LoveDA validation set using different α values. The highest score is shown in bold.

| α | Methods | background | building | road | water | barren | forest | agricultural | mIoU |
|---|---|---|---|---|---|---|---|---|---|
| | C | 55.58 | 66.08 | 56.32 | 68.61 | 32.82 | 42.33 | **55.88** | 53.95 |
| N/A | V | 55.29 | 65.08 | 56.89 | 71.31 | 33.68 | 40.17 | 53.52 | 53.71 |
| | S | 54.61 | 65.37 | 55.98 | 69.79 | 31.69 | 44.14 | 52.19 | 53.40 |
| | C+V | 56.20 | 67.01 | 57.91 | 72.00 | 34.19 | 41.26 | 55.30 | 54.84 |
| 1.0 | C+S | 55.85 | 67.17 | 57.16 | 71.06 | 32.96 | 43.52 | 54.77 | 54.64 |
| | S+V | 55.69 | 66.90 | 57.82 | 71.78 | 33.58 | 41.89 | 53.55 | 54.46 |
| | C+V | 56.23 | 66.86 | 57.84 | **72.10** | 34.04 | 41.24 | 55.36 | 54.81 |
| 1.5 | C+S | 55.93 | 67.25 | 57.23 | 71.12 | 32.89 | 43.60 | 54.83 | 54.69 |
| | S+V | 55.70 | 67.17 | 57.81 | 71.70 | 33.79 | 42.09 | 53.43 | 54.53 |
| 2.0 | C+V | 56.19 | 66.86 | 58.01 | 71.98 | 34.18 | 41.26 | 55.42 | 54.84 |
| | C+S | 55.90 | 67.41 | 57.25 | 70.96 | 33.03 | 43.66 | 54.83 | 54.72 |
| | S+V | 55.68 | 67.14 | 57.85 | 71.69 | 33.83 | 42.10 | 53.45 | 54.53 |
| | C+V | **56.25** | 67.06 | **57.91** | 72.02 | 34.17 | 41.39 | 55.42 | **54.89** |
| 2.5 | C+S | 55.94 | **67.53** | 57.24 | 70.91 | 32.99 | 43.71 | 54.84 | 54.74 |
| | S+V | 55.61 | 66.78 | 57.81 | 71.68 | 33.80 | 41.94 | 53.53 | 54.45 |
| | C+V | 56.23 | 67.25 | 57.78 | 71.94 | **34.23** | 41.47 | 55.35 | 54.89 |
| 3.0 | C+S | 55.87 | 67.48 | 57.12 | 70.75 | 33.07 | **43.74** | 54.80 | 54.69 |
| | S+V | 55.71 | 66.94 | 57.72 | 71.77 | 33.73 | 42.15 | 53.54 | 54.51 |

**Table 9.** Comparison of segmentation performance on the Vaihingen validation set using different α values. The highest score is shown in bold.

| α | Methods | surface | building | vegetation | tree | car | clutter | mIoU |
|---|---|---|---|---|---|---|---|---|
| | C | 83.35 | 90.65 | 68.75 | 82.63 | 72.57 | 49.68 | 79.59 |
| N/A | V | 83.39 | 91.11 | 68.91 | 83.04 | 70.46 | 49.51 | 79.38 |
| | S | 84.00 | 91.16 | 68.55 | 82.67 | 72.03 | 50.36 | 79.68 |
| | C+V | 84.29 | 91.18 | 70.02 | 83.51 | 73.42 | 50.07 | 80.48 |
| 1.0 | C+S | **84.45** | 91.43 | 69.92 | 83.37 | 73.14 | 50.04 | 80.46 |
| | S+V | 84.36 | 91.54 | **70.11** | 83.59 | 72.02 | 50.39 | 80.32 |
| | C+V | 84.30 | 91.21 | 70.03 | 83.51 | 73.21 | 50.06 | 80.45 |
| 1.5 | C+S | 84.41 | 91.41 | 69.98 | 83.41 | 72.49 | 50.08 | 80.34 |
| | S+V | 84.34 | 91.51 | 70.05 | 83.53 | 72.19 | 50.36 | 80.32 |
| | C+V | 84.29 | 91.19 | 70.05 | 83.50 | 73.13 | 50.07 | 80.43 |
| 2.0 | C+S | 84.44 | 91.41 | 69.91 | 83.36 | 73.20 | 50.06 | 80.46 |
| | S+V | 84.38 | 91.56 | 70.03 | 83.52 | 72.35 | **50.39** | 80.37 |
| | C+V | 84.29 | 91.18 | 69.94 | 83.46 | **73.83** | 50.08 | **80.54** |
| 2.5 | C+S | 84.41 | 91.42 | 69.87 | 83.34 | 73.27 | 50.06 | 80.46 |
| | S+V | 84.36 | 91.48 | 69.89 | 83.42 | 73.35 | 50.38 | 80.50 |
| | C+V | 84.29 | 91.21 | 70.06 | 83.53 | 73.07 | 50.07 | 80.43 |
| 3.0 | C+S | 84.44 | 91.43 | 69.94 | 83.37 | 73.10 | 50.05 | 80.46 |
| | S+V | 84.39 | **91.54** | 70.10 | **83.58** | 72.23 | 50.37 | 80.37 |

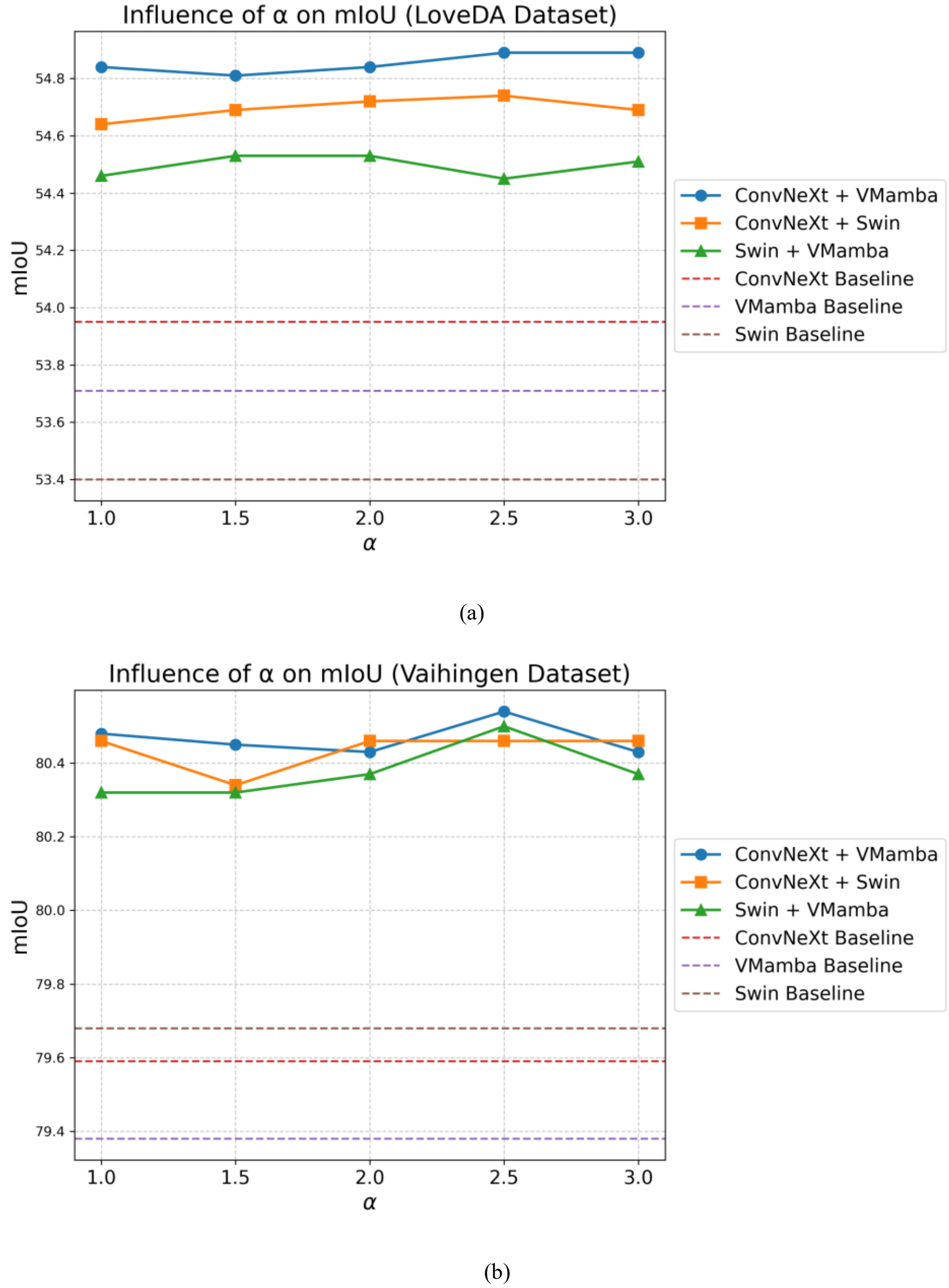

**Fig 3.** Line chart of the segmentation performance using different α values. (1) LoveDA validation set, (2) Vaihingen validation set.

### 4.4.3 Influence of number of expert networks

To investigate the impact of varying the number of selected expert networks ($K$) on segmentation performance and address concerns about potential saturation or redundancy with larger $K$, we conducted additional experiments on the LoveDA and Vaihingen validation sets. The greedy algorithm (Section 3.1) was used to select networks progressively, with α fixed at 2.5 for consistency. For LoveDA, the selections are: $K = 2$ (C+V), $K = 3$ (C+S+V), $K = 4$ (C+S+V+Segformer). For Vaihingen, they are: $K = 2$ (C+S), $K = 3$ (C+V+S), $K = 4$ (C+S+V+BiseNet).

**Table 10.** Ablation on varying $K$ values for segmentation performance on LoveDA validation sets. The highest score is shown in bold.

| $K$ | Methods | background | building | road | water | barren | forest | agricultural | mIoU |
|---|---|---|---|---|---|---|---|---|---|
| 2 | C+V | **56.25** | 67.06 | **57.91** | 72.02 | **34.17** | 41.39 | **55.42** | 54.89 |
| 3 | C+V+S | 56.14 | 67.64 | 57.84 | 72.11 | 33.76 | 42.62 | 54.58 | **54.96** |
| 4 | C+V+S+Segformer | 55.99 | **68.00** | 57.68 | **72.56** | 33.34 | **42.87** | 53.28 | 54.82 |

**Table 11.** Ablation on varying $K$ values for segmentation performance on Vaihingen validation sets. The highest score is shown in bold.

| $K$ | Methods | surface | building | vegetation | tree | car | clutter | mIoU |
|---|---|---|---|---|---|---|---|---|
| 2 | C+S | 84.41 | 91.42 | 69.87 | 83.34 | **73.27** | 50.06 | 80.46 |
| 3 | C+V+S | 84.53 | 91.53 | 70.23 | 83.53 | 72.93 | 50.24 | **80.55** |
| 4 | C+V+S+BiSeNet | **84.62** | **91.57** | **70.35** | **83.60** | 71.38 | **50.37** | 80.30 |

As K increased, mIoU improved initially due to greater expert diversity but showed diminishing returns or slight declines for larger $K$, as shown in **Table 10** and **Table 11**. For instance, at $K = 4$, performance was enhanced in specific categories like building (68.00 > 67.64), water (72.56 > 72.11), and forest (42.87 > 42.62) on LoveDA, and all categories except car on Vaihingen, but this did not necessarily translate to higher overall mIoU (e.g., 54.82 < 54.96 on LoveDA; 80.30 < 80.55 on Vaihingen). This occured because incorporating additional networks like Segformer or BiseNet may introduce suboptimal predictions that diluted the fused results through exponential weighting, especially in categories with high variance among experts. Therefore, in practice, we do not recommend setting $K$ too large, as it may lead to redundancy and computational overhead without proportional gains. Instead, we found that $K = 3$ often led to a balance for efficiency and performance on these datasets.

### 4.4.4 Validity of category-specific fusion

The ClassWise-CRF architecture ingeniously combines category-specific fusion with CRF optimization, significantly enhancing semantic segmentation performance. To validate the effectiveness of category-specific fusion in our approach, we independently applied CRF optimization to three baseline methods and compared their results with those of the ClassWise-CRF architecture. The experimental outcomes were evaluated on the LoveDA and Vaihingen validation sets, with specific data presented in **Table 12** and **Table 13**, respectively. The results indicate that applying CRF optimization alone to the baseline methods yields minimal improvements in segmentation performance, and in some cases, even leads to performance degradation. In contrast, the ClassWise-CRF architecture, by organically integrating category-specific fusion with CRF optimization, achieves substantial performance gains.

**Table 12.** Comparison of the segmentation performance of the CRF strategy with and without the fusion method on the LoveDA validation set. The highest score is shown in bold.

| Strategy | Methods | background | building | road | water | barren | forest | agricultral | mIoU |
|---|---|---|---|---|---|---|---|---|---|
| Baseline | C | 55.58 | 66.08 | 56.32 | 68.61 | 32.82 | 42.33 | 55.88 | 53.95 |
| | V | 55.29 | 65.08 | 56.89 | 71.31 | 33.68 | 40.17 | 53.52 | 53.71 |
| | S | 54.61 | 65.37 | 55.98 | 69.79 | 31.69 | 44.14 | 52.19 | 53.40 |
| w/o fusion | C | 55.80 | 65.76 | 56.75 | 68.93 | 32.61 | 42.07 | **56.35** | 54.04 |
| | V | 55.47 | 64.81 | 57.40 | 71.56 | 33.62 | 40.00 | 53.90 | 53.82 |
| | S | 54.80 | 65.36 | 56.25 | 70.01 | 31.48 | 43.96 | 52.52 | 53.48 |
| ClassWise-CRF | C+V | **56.25** | 67.06 | **57.91** | **72.02** | **34.17** | 41.39 | 55.42 | **54.89** |
| | C+S | 55.94 | **67.53** | 57.24 | 70.91 | 32.99 | **43.71** | 54.84 | 54.74 |
| | S+V | 55.61 | 66.78 | 57.81 | 71.68 | 33.80 | 41.94 | 53.53 | 54.45 |

**Table 13.** Comparison of the segmentation performance of the CRF strategy with and without the fusion method on the Vaihingen validation set. The highest score is shown in bold.

| Strategy | Methods | surface | building | vegetation | tree | car | clutter | mIoU |
|---|---|---|---|---|---|---|---|---|
| Baseline | C | 83.35 | 90.65 | 68.75 | 82.63 | 72.57 | 49.68 | 79.59 |
| | V | 83.39 | 91.11 | 68.91 | 83.04 | 70.46 | 49.51 | 79.38 |
| | S | 84.00 | 91.16 | 68.55 | 82.67 | 72.03 | 50.36 | 79.68 |
| w/o fusion | C | 84.02 | 90.78 | 69.31 | 82.94 | 72.04 | 49.80 | 79.82 |
| | V | 84.14 | 91.24 | 68.94 | 82.88 | 71.74 | **50.40** | 79.79 |
| | S | 83.40 | 91.13 | 69.25 | 83.23 | 69.26 | 49.96 | 79.25 |
| ClassWise-CRF | C+V | 84.29 | 91.18 | **69.94** | **83.46** | **73.83** | 50.08 | **80.54** |
| | C+S | **84.41** | 91.42 | 69.87 | 83.34 | 73.27 | 50.06 | 80.46 |
| | S+V | 84.36 | **91.48** | 69.89 | 83.42 | 73.35 | 50.38 | 80.50 |

## 4.4.5 Time cost of network selection methods

As discussed in Section 3.1, when the pool of candidate networks ($N$) is small, the time costs of our proposed greedy algorithm and the exhaustive search differ minimally, making both viable options. However, when the pool of candidate networks ($N$) is large, our greedy algorithm offers a clear efficiency advantage. We simulated IoU matrices with randomly generated values between 0 and 100, assuming $C = 7$ categories (consistent with the LoveDA dataset). Experiments were conducted on a standard CPU environment (AMD Ryzen 7 3700X 8-Core Processor × 16, Python 3.9) without GPU acceleration. The greedy algorithm exhibits a time complexity of $O(K \cdot N \cdot C)$, as it iterates $K$ times, scanning up to $N$ candidates per iteration and computing the mIoU across $C$ categories. In contrast, the exhaustive search has a time complexity of $O\left(\binom{N}{K} \cdot K \cdot C\right)$, where $\binom{N}{K}$ denotes the binomial coefficient representing the number of combinations, rendering it computationally prohibitive for large $N$ and $K$ due to its exponential scaling.

The results are summarized in **Table 14**. For $N = 10$, both methods were highly efficient, with time differences under 0.002 seconds. However, for $N = 100$, the exhaustive search scaled poorly, requiring over 34 seconds for $K = 4$, whereas the greedy algorithm remained under 0.004 seconds across all $K$ values. These findings affirm the greedy approach's suitability for large candidate pools, as it delivers near-optimal selections with negligible overhead. As noted in Section 3.1, network selection is a one-time procedure during the deployment phase and does not affect real-time inference efficiency.

**Table 14.** Time cost (in seconds) of network selection algorithms for different $N$ and $K$.

| $N$ | $K$ | Greedy Time (s) | Exhaustive Time (s) |
|---|---|---|---|
| | 2 | 0.0003 | 0.0004 |
| 10 | 3 | 0.0003 | 0.0011 |
| | 4 | 0.0003 | 0.0019 |
| | 2 | 0.0018 | 0.0434 |
| 10 | 3 | 0.0028 | 1.4512 |
| | 4 | 0.0033 | 34.4892 |

## 5 Discussion

The ClassWise-CRF architecture is fundamentally a result-level fusion approach, with its performance contingent upon the prediction outputs of the baseline networks. Consequently, in practical applications, the inference time of this architecture typically exceeds the cumulative inference times of the individual baseline networks. Specifically, the total inference time $t_{\text{total}}$ can be expressed as:

$$t_{\text{total}} = \sum_{i=1}^{K} t_{\text{net}_i} + t_{\text{fusion}} \quad (14)$$

where $t_{\text{net}_i}$ represents the inference time of the $i$-th baseline network, and $t_{\text{fusion}}$ denotes the time required for the fusion process. In our experiments, conducted on a single NVIDIA RTX 3090 GPU, the fusion process—including probability map fusion and CRF optimization—takes approximately 0.5 seconds per image, compared to about 0.2 seconds per image for a single baseline model, highlighting a moderate trade-off where the added overhead yielded measurable accuracy gains but might constrain scalability in high-throughput scenarios. This characteristic renders ClassWise-CRF more suitable for scenarios with relaxed real-time requirements, such as semantic segmentation analysis of remote sensing imagery [4]. In applications demanding stringent real-time performance [65], such as autonomous driving [66, 67], the strict constraints on response latency may limit the practicality of integrating multiple baseline networks. Additionally, the Bayesian optimization process for hyperparameter tuning, performed on the validation set, introduces further computational overhead. In the Vaihingen dataset, each optimization trial takes approximately 36 seconds on an NVIDIA RTX 3090 GPU. With a setting of 20 trials, the total optimization time amounts to about 721 seconds. Although this optimization is a one-time cost during the model development phase, it is an important consideration when adapting the architecture to new datasets or tasks. Nevertheless, with continuous advancements in modern GPU technology, the marked improvements in network inference speed open new avenues for the potential deployment of ClassWise-CRF across a wider array of semantic segmentation tasks. Future research could focus on developing more efficient optimization techniques to reduce $t_{\text{fusion}}$, thereby enhancing the architecture's applicability in time-sensitive scenarios.

Furthermore, as elaborated in Section 4.3.2, although the method exhibits low sensitivity to the hyperparameter α (which governs the degree of weight amplification), the CRF optimization process involves multiple parameters that necessitate systematic tuning via Bayesian optimization on the validation set to achieve an optimal configuration. While this tuning is effective, it also elevates computational complexity. Our approach relies on validation set metrics as priors for category-specific weighting, which assumes access to sufficient labeled validation data, a common setup in fully supervised benchmarks like LoveDA and Vaihingen. However, in real-world scenarios with limited labeled validation data (e.g., semi-supervised or few-shot settings), this reliance could limit generalization. Beyond remote sensing datasets, the ClassWise-CRF framework could be extended to semantic segmentation of UAV-captured images [68] and medical images [69] in the future, enhancing related applications. UAV imagery often features multi-scale objects and dynamic environmental conditions akin to satellite-based remote sensing, where our category-specific fusion can boost segmentation precision for tasks like aerial monitoring. Similarly, medical image segmentation shares similarities with remote sensing in handling class imbalances and requiring fine boundary delineation for organs or tissues, suggesting the potential applicability of our framework to improve diagnostic and analytical outcomes in healthcare. Building on the preceding discussion and findings, we propose the following directions for future research:

1. Investigate optimization techniques that can supplant CRF to deliver adaptive and robust optimization outcomes, thereby improving segmentation accuracy while reducing computational overhead.

2. Extend the fusion framework to object detection domains by synthesizing prediction results from multiple networks to further enhance detection performance, while exploring optimization strategies (such as lightweight fusion variants) to ensure inference speed meets real-time requirements in time-sensitive applications.

3. Employ model distillation techniques to consolidate knowledge from networks proficient in distinct categories, enabling the distilled model to harness their collective strengths.

4. Explore alternative prior estimation techniques, such as self-supervised metrics derived from unlabeled data or pseudo-labeling from ensemble predictions, to adapt ClassWise-CRF to data-scarce environments without compromising its fusion efficacy.

5. To address challenges with minority or underrepresented categories, develop category-specific resampling to balance class distributions during expert selection or adaptive priors that dynamically adjust weights based on category difficulty, thereby improving fusion efficacy for imbalanced datasets.

# 6 Conclusion

This paper introduces ClassWise-CRF, a novel result-level fusion architecture for semantic segmentation, which markedly elevates the accuracy and robustness of remote sensing image segmentation through category-specific fusion and CRF optimization. The core innovation of ClassWise-CRF lies in its ability to adaptively fuse the superior prediction results of multiple networks across different categories. By leveraging a network selection algorithm, the method efficiently identifies top-performing expert networks for individual semantic categories from a pool of candidate networks and fuses their prediction probability maps using an exponential weighting strategy. The fused outcomes are subsequently refined via CRF optimization, which enhances spatial consistency and boundary accuracy.

Experimental results show that the performance of ClassWise-CRF surpasses baseline methods in terms of mIoU and most IoU metrics on the LoveDA and Vaihingen datasets. Ablation studies further validate the effectiveness of category-specific fusion, underscoring the method's breakthrough in advancing segmentation performance. Through this study, we have achieved consistent mIoU improvements of 0.68%-1.00% over strong baselines, demonstrating the framework's ability to harness complementary network strengths for enhanced segmentation accuracy in challenging remote sensing scenarios, while maintaining computational feasibility for deployment.

This work contributes to academics by introducing a plug-and-play, category-specific fusion paradigm that addresses performance variability across semantic classes, enriching the theoretical landscape of ensemble methods in deep learning-based segmentation. In practice, it offers tangible benefits for remote sensing applications, such as improved land cover mapping for urban planning (where enhanced accuracy in segmenting buildings and roads reduces misclassification errors, leading to precise land-use maps and minimized costly revisions), agricultural monitoring (improving delineation of barren, forest, and agricultural areas for accurate crop yield estimation and efficient precision farming), and disaster assessment (enabling finer segmentation of water bodies and vegetation for faster flood risk evaluation and timely response planning).

Future research can explore efficient alternatives to CRF optimization, extensions to object detection and data-scarce environments, model distillation for lightweight deployment, and strategies to better handle minority categories in imbalanced datasets.

## Declaration of competing interest

The authors declare that they have no known competing financial interests or personal relationships that could have appeared to influence the work reported in this paper.

## Acknowledgments

This work was supported in part by the Xi'an Jiaotong-Liverpool University Research Enhancement Fund under Grant REF-21-01-003, and in part by the Xi'an Jiaotong-Liverpool University Postgraduate Research Scholarship under Grant FOS2210JJ03.